\documentclass[10pt,twocolumn,letterpaper]{article}

\usepackage{cvpr}              
\definecolor{cvprblue}{rgb}{0.21,0.49,0.74}
\usepackage[pagebackref,breaklinks,colorlinks,allcolors=cvprblue]{hyperref}

\def\paperID{} 
\def\confName{CVPR}
\def\confYear{2026}

\title{UniGenDet: A Unified Generative-Discriminative Framework for Co-Evolutionary Image Generation and Generated Image Detection
}

\author{
Yanran Zhang~
Wenzhao Zheng$^\dagger$~
Yifei Li~
Bingyao Yu~
Yu Zheng~
Lei Chen~
Jiwen Lu~
Jie Zhou$^{*}$\\
Department of Automation, Tsinghua University, China\\
Project Page: {\tt \href{https://ivg-yanranzhang.github.io/UniGenDet/}{https://ivg-yanranzhang.github.io/UniGenDet/} }
}

\begin{document}
\maketitle


\begin{abstract}
In recent years, significant progress has been made in both image generation and generated image detection. Despite their rapid, yet largely independent, development, these two fields have evolved distinct architectural paradigms: the former predominantly relies on generative networks, while the latter favors discriminative frameworks. A recent trend in both domains is the use of adversarial information to enhance performance, revealing potential for synergy. However, the significant architectural divergence between them presents considerable challenges. Departing from previous approaches, we propose \textbf{UniGenDet:} a \textbf{Uni}fied generative-discriminative framework for
co-evolutionary image \textbf{Gen}eration and generated image \textbf{Det}ection. To bridge the task gap, we design a symbiotic multimodal self-attention mechanism and a unified fine-tuning algorithm. This synergy allows the generation task to improve the interpretability of authenticity identification, while authenticity criteria guide the creation of higher-fidelity images. Furthermore, we introduce a detector-informed generative alignment mechanism to facilitate seamless information exchange. Extensive experiments on multiple datasets demonstrate that our method achieves state-of-the-art performance. Code: \href{https://github.com/Zhangyr2022/UniGenDet}{https://github.com/Zhangyr2022/UniGenDet}.
\end{abstract}    
\section{Introduction}
\begin{figure}[t]
  \centering
  \includegraphics[width=\linewidth]{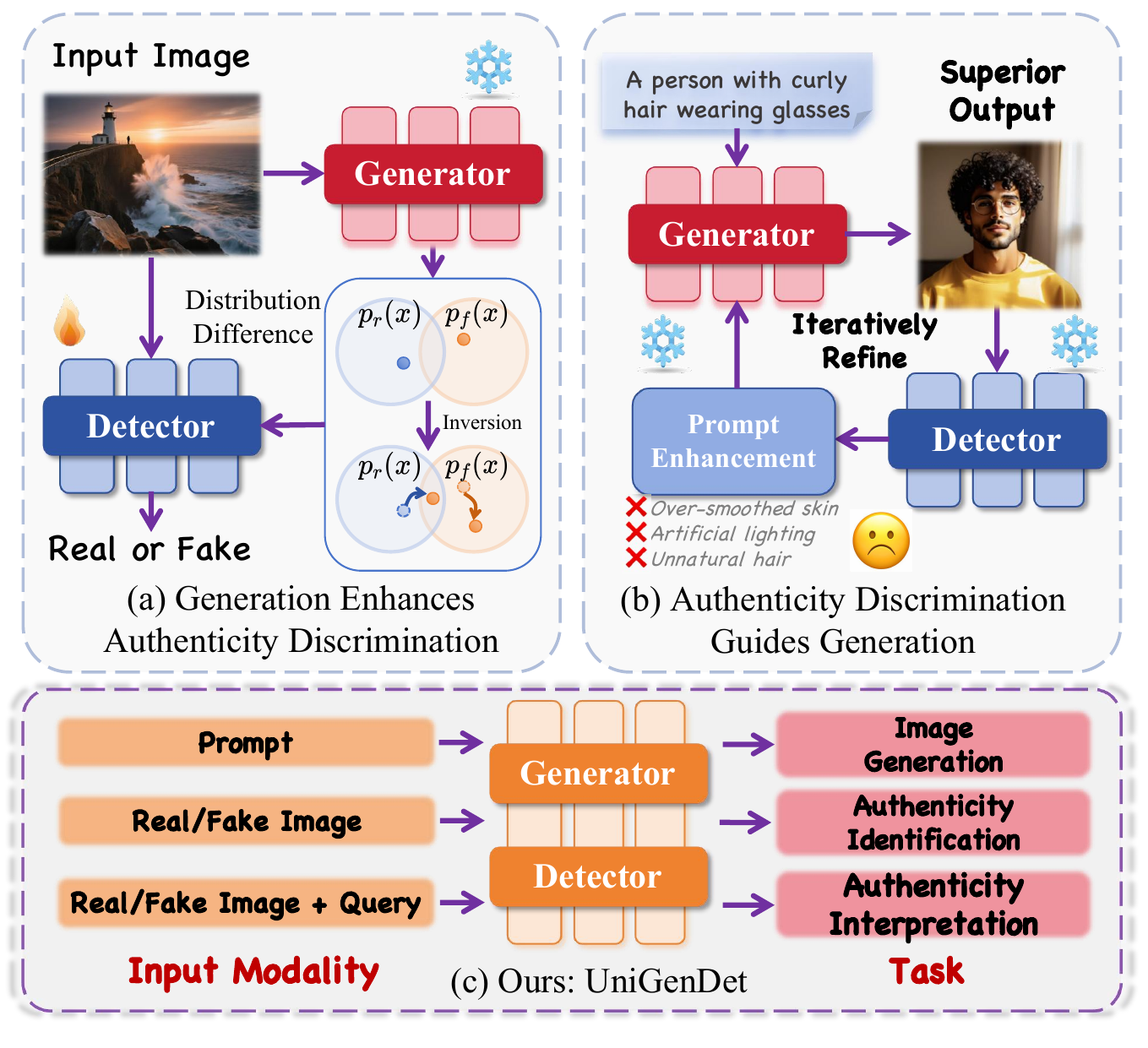}
  \vspace{-9mm}
  \caption{Our unified framework bridges generation and authenticity discrimination synergistically. 
(a) Generation enhances detection by reducing distributional gaps. 
(b) Detection feedback refines generation for higher realism. 
(c) Our unified model supports multiple input modalities and tasks within a single architecture.}
   \label{fig:teaser}
   \vspace{-7mm}
\end{figure}

The “spear” of generative AI and the “shield” of generated content authentication are in a constant arms race, driving each other’s evolution with significant societal impact. Generative technology has matured, marked by parallel advances in GANs~\cite{jeong2022frepgan}, VAEs~\cite{mentzer2023finite}, diffusion models~\cite{zhang2025ICEdit,qwen-image}, and autoregressive models~\cite{tian2024VAR,Sun24LlamaGen}. Recent milestones, such as the Diffusion Transformer and high-fidelity models like GPT-4o~\cite{achiam2023gpt}, Sora~\cite{openai2025sora}, Nano Banana Pro~\cite{nanobanana_pro}, are now accessible to the masses. Tools like ROOP v3.0 allow for 10-second face-swapping on a phone, fueling both creative applications and a dark side: a black market where counterfeit content has surged 1000-fold in two years~\cite{runway2025gen4}.
Generated image detection technologies are advancing through visual feature analysis and content plausibility checks~\cite{huang2024sida,liu2024forgery}. Solutions can now pinpoint pixel-level artifacts and lighting inconsistencies, with multimodal fusion enabling fast verification across diverse media.
However, most detection pipelines are still trained \emph{in isolation} on snapshots of existing generators, treating forgeries as a moving target rather than a co-evolving process. As generative models rapidly update their architectures and post-processing, detectors often overfit to transient cues, suffer from domain gaps to unseen generators, and lack access to the generative logic behind forgeries—making detection capabilities struggle to keep pace with the sophistication of new forgery methods.

Generation and detection are advancing in parallel. As architectures evolve, generators begin to integrate auxiliary signals by leveraging understanding models for prior knowledge, hinting at a unified generation–understanding paradigm. Meanwhile, detection is moving toward large models with stronger generalization to novel attacks and improved interpretability. However, as shown in Figure~\ref{fig:teaser}, existing methods largely optimize generation and detection separately: 
generators pursue perceptual realism, yet still manifest discernible artifacts like physical inconsistencies, without being explicitly constrained by forensic requirements,
while detectors reactively learn from fixed or lagging forgeries. Despite this convergence, a closed-loop framework that \emph{co-optimizes} both remains absent.

As Richard Feynman said, “\textit{What I cannot create, I do not understand}”, which profoundly reveals the essential symbiotic relationship between generation and discrimination. Motivated by this, we propose, to our knowledge, a unified generative-discriminative framework for co-evolutionary image generation and generated image detection for the first time.
To further enhance this synergy, we introduce a generation-detection unified fine-tuning algorithm, equipping the generated image detection model with an understanding of the generative logic of forgeries. This allows it to more effectively discern the core of the truth-fake boundary, thereby improving generalization and interpretability. Moreover, we adopt detector-informed generative alignment, forming a feedback loop in which the generative process is refined by the rigorous demands of generated image detection, compelling it to better approximate the underlying rules of reality.
\label{sec:intro}

\section{Related Work}

\noindent\textbf{Synthetic Image Detection and Explanation.}
Research in synthetic image detection has evolved from binary classification to explainable analysis. Early specialist models~\cite{mccloskey2018detecting,wang2020cnn,frank2020leveraging,liu2022detecting,tan2024rethinking,wang2023dire}, targeting artifacts from specific generators, achieved high accuracy on many benchmarks~\cite{wang2020cnn,ojha2023towards,zhu2023genimage, corvi2023detection, zhang2025d3qe} but suffered from poor generalization and lacked interpretability. The advent of Large Multimodal Models (LMMs)~\cite{li2024llava,google2025gemini,openai2025chatgpt4o,qwen2.5-vl} introduced joint detection and explanation capabilities. Works like LOKI~\cite{ye2024loki} and Fakebench~\cite{li2024fakebench} demonstrated that LMMs could provide natural language justifications for their decisions, enhancing transparency. However, the classification accuracy of pretrained models often lags behind specialist models. Consequently, recent efforts have explored fine-tuning LMMs on detection datasets (e.g., SIDA~\cite{huang2024sida}, FakeVLM~\cite{wen2025spot}, Legion~\cite{kang2025legion}, Skyra~\cite{li2025skyra}) or developing hybrid architectures. These hybrid systems, such as ForgeryGPT~\cite{li2024forgerygpt}, X2-DFD~\cite{chen2024x2}, FFAA~\cite{huang2024ffaa}, and AIGIHolmes~\cite{zhou2025aigi}, fuse the discriminative power of traditional detectors with the reasoning of LMMs. This approach seeks to combine high accuracy with interpretability, though generalization and robustness remain open challenges. Moreover, current detection methods consistently lag behind emerging generative models, creating a critical gap in timely synthetic media detection. Studying the co-evolution of generation and detection is therefore essential for achieving strong generalization and long-term effectiveness.

\noindent\textbf{Unified Visual Generation and Understanding.}
Unified models for visual generation and understanding seek to develop a single architecture that handles both multimodal comprehension and synthesis, overcoming the limitations of specialized models. Historically, understanding tasks were dominated by autoregressive models~\cite{li2024llava,qwen2.5-vl}, while generation relied on separate diffusion~\cite{ddpm,rombach2022high,flux,SD3} or autoregressive~\cite{EsserRO21VQGAN,Sun24LlamaGen,tian2024VAR} frameworks. Recent advancements, spurred by systems like GPT-4o~\cite{openai2025chatgpt4o}, have shifted focus toward unified encoder-decoder architectures that frame both tasks as a sequence modeling problem. Current approaches primarily fall into three categories~\cite{zhang2025unified}: diffusion-based models that employ dual processes for joint generation~\cite{li2025dual,yang2025mmada}, autoregressive models that leverage multi-scale visual tokenizers~\cite{deng2025emerging,liao2025mogao,xie2024show}, and hybrid architectures that merge autoregressive reasoning with diffusion-based synthesis~\cite{emu3,chameleon,chen2025blip3}. Recent works utilize techniques such as Mixture-of-Transformer~\cite{deng2025emerging}, shared masked autoencoders~\cite{wu2025harmonizing}, or dual-codebook structures to harmonize representations for both tasks~\cite{qu2024tokenflow}, achieving performance gains and consistency through modality alignment mechanisms like semantic encoders or pixel-level tokenizers.

\noindent\textbf{Co-evolution of Generation and Synthetic Image Detection.}
The synergy between content generation and detection is crucial for a trustworthy AI ecosystem, yet their deep interaction within unified Multimodal Large Language Models (MLLMs) remains under-explored. AI-generated image detection, a specialized understanding task focused on local inconsistencies, can provide valuable feedback to improve generation quality. Conversely, a generator's knowledge of the synthetic data distribution can enhance a detector's sensitivity. While the adversarial dynamic in GANs exemplifies this synergy, its training paradigm is ill-suited for large-scale MLLMs.
Some works have leveraged generative knowledge for detection. DIRE~\cite{wang2023dire} pioneered using a diffusion model's reconstruction error to spot its own outputs, a concept extended by AEROBLADE~\cite{ricker2024aeroblade} with autoencoders and LARE$^2$~\cite{luo2024lare} with latent representation errors. However, these methods are often model-specific and lack interpretability.
In the reverse direction, detection can guide generation. The LEGION~\cite{kang2025legion} framework, for instance, uses a discriminator during inference to evaluate and iteratively refine generated images by optimizing prompts. Therefore, designing 
efficient mechanisms for co-evolution between generation and detection within a unified framework is a critical and promising research direction.

\begin{figure*}[t]
  \centering
  \includegraphics[width=\linewidth]{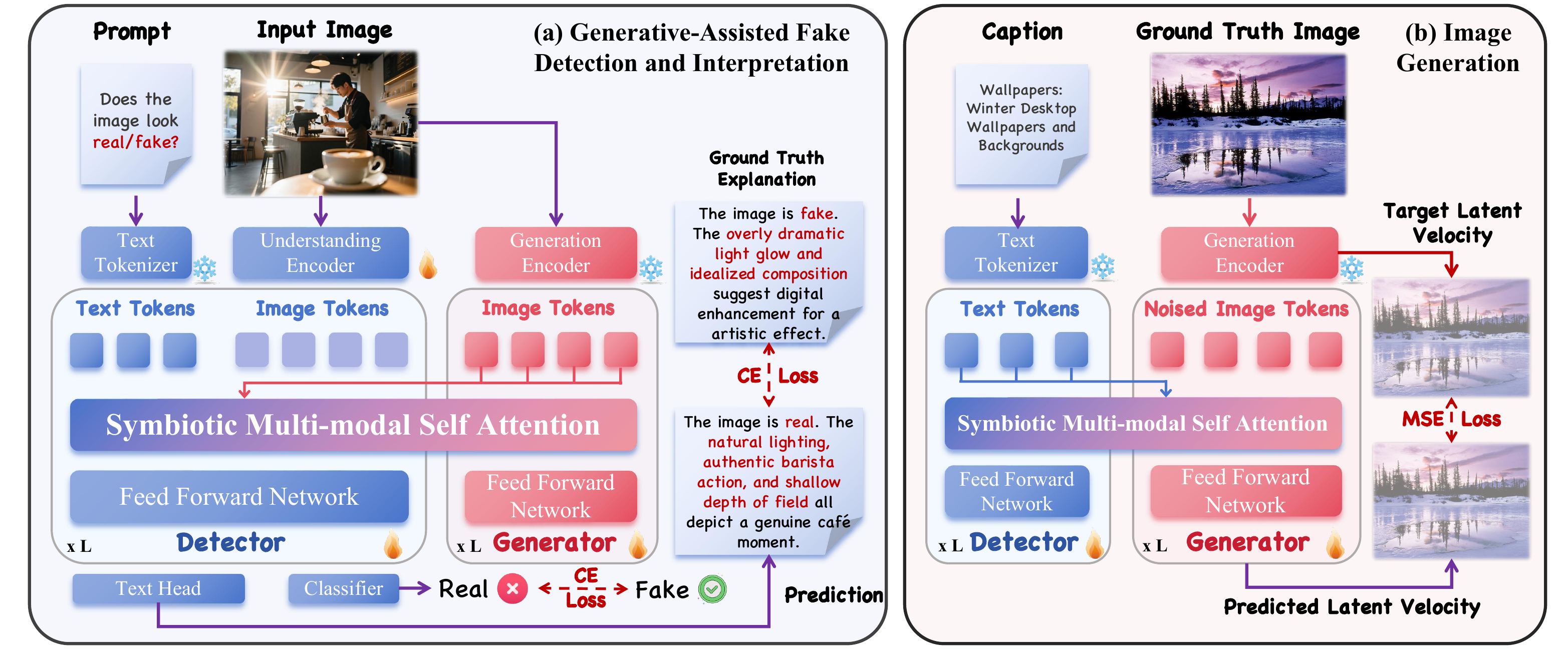}
  \vspace{-7mm}
  \caption{{\textbf{Overview of the Generation–Detection Unified Fine-tuning (GDUF) pipeline.} 
(a) \textit{Generative-Assisted Fake Detection and Interpretation:} the \textbf{Symbiotic Multi-modal Self-Attention (SMSA)} guides the detector using generator features for authenticity analysis and textual explanation. 
(b) \textit{Image Generation:} discriminative cues from the detector are injected into the generator for authenticity-aware synthesis. 
{\textcolor{red}{\faFire}}~denotes trainable modules, and {\textcolor{blue}{\faSnowflake}}~indicates frozen ones.
}}
   \label{fig:pipeline}
  \vspace{-3mm}
\end{figure*}

\section{Approach}

In this section, we systematically introduce the \textbf{UniGenDet} framework, which employs a two-stage training paradigm to synergistically optimize AI-generated image detection, explanation, and image generation. The first stage performs task-adaptive fine-tuning of a unified generation-detection model, while the second stage leverages the trained detector to adversarially optimize the generator.

\subsection{Problem Formulation}
Our framework simultaneously addresses AI-generated image detection with explanation and image generation. For detection, given an input image $\mathcal{I} \in \mathbb{R}^{H\times W\times 3}$ and textual instruction $\mathcal{Q}$, the model outputs a binary label ${y} \in \{\text{Real}, \text{Fake}\}$ and semantic explanation $\mathcal{A}$ articulating the decision basis through visual features like generation artifacts or semantic inconsistencies. For generation, given prompt $\mathcal{P}$, the model produces $\hat{\mathcal{I}} \in \mathbb{R}^{H\times W\times 3}$ that is semantically consistent and visually realistic. This unified framework enables co-evolution of multimodal understanding and generation capabilities, providing a closed-loop solution for generative model security evaluation.

\subsection{Generation-Detection Unified Fine-tuning }

\noindent\textbf{Design Insights.} Previous detection methods~\cite{ye2024loki,yan2024sanity,kang2025legion} demonstrate that detecting AI-generated images remains challenging amid rapid generative model evolution. While traditional methods focus on binary classification, recent works incorporate Vision-Language Models (e.g., LLaVA~\cite{li2024llava}) to achieve detection with explanation, enhancing interpretability and generalization. However, some studies~\cite{li2024forgerygpt,huang2024sida,kang2025legion,zhou2025aigi,wen2025spot} show that vision-language models fine-tuned on specific datasets exhibit limited zero-shot cross-dataset performance, struggling to adapt to evolving generative models. The fundamental issue is that detection development lags behind generation advances, creating reactive defense.

Recent research reveals bidirectional benefits between generation and detection: DIRE~\cite{wang2023dire} and LARE$^2$~\cite{luo2024lare} confirm that generative distributional features enhance detection accuracy, while LEGION~\cite{kang2025legion} shows detection signals optimize generation quality. However, these optimizations remain unidirectional and static. This motivates our unified framework with three advantages: (1) \textbf{Task Synergy}: Detection and generation mutually improve performance, with the detector more accurately identifying forgery traces and the generator improving its output quality through this feedback; (2) \textbf{Synchronized Evolution}: The unified framework enables simultaneous evolution of detection and generation, mitigating the traditional ``detection lag" to enhance generalization and timeliness; (3) \textbf{System Efficiency}: shared parameters reduce deployment costs while end-to-end training enhances knowledge transfer. These synergistic advantages establish a foundational rationale for our unified approach. We now elaborate on the model architecture that materializes this design.

\noindent\textbf{Model Architecture.} We select BAGEL~\cite{deng2025emerging} as our base model, which employs a multimodal Mixture of Transformers architecture supporting both image generation and visual question answering. Its multi-task compatibility provides a solid foundation for our approach. As illustrated in Figure ~\ref{fig:pipeline}, we perform joint fine-tuning using existing detection and generation datasets, placing generation and detection under a unified training objective to enable synergistic optimization within a model.

For the \textit{detection task}, each training sample contains input image $\mathcal{I}$, a text instruction $\mathcal{Q}$ (e.g., ``Does the image look real/fake?"), and answer $\mathcal{A}$ aggregated from multi-model annotations. Considering the strong modeling capability of generative models on image distributions, particularly in the latent space of diffusion models where semantic and structural information is well-preserved, we design a \textbf{Symbiotic Multi-modal Self-Attention (SMSA)} module to facilitate knowledge transfer from generation to detection. Specifically, the input image is processed through both the detection image encoder (SigLIP~\cite{tschannen2025siglip}) and generation image encoder (FLUX VAE encoder~\cite{flux}), yielding detection features $h_{\text{det}}^{(0)} = \mathcal{E}_{\text{det}}(\mathcal{I})$ and generation latents $z_{\text{gen}}^{(0)} = \mathcal{E}_{\text{gen}}(\mathcal{I})$, respectively. Meanwhile, the text instruction $\mathcal{Q}$ is encoded into textual embeddings $h_{\text{text}}^{(0)}$. 
The SMSA module is applied at each layer $l$ of the detector backbone, where generation latents interact with detection features and textual representations through cross-modal attention. Specifically, we first concatenate the three modality features:
\begin{equation}
    h_{\text{concat}}^{(l)} = [z_{\text{gen}}^{(l)}; h_{\text{det}}^{(l)}; h_{\text{text}}^{(l)};].
\end{equation}
Then, multi-head cross-attention is computed as:
\begin{equation}
    \text{Attention}(Q, K, V) = \text{softmax}\left(\frac{QK^\top}{\sqrt{d_k}}\right)V,
\end{equation}
where $Q = W_Q h_{\text{det}}^{(l)}$, $K = W_K h_{\text{concat}}^{(l)}$, $V = W_V h_{\text{concat}}^{(l)}$, and $W_Q, W_K, W_V$ are learnable projection matrices. The multi-head attention aggregates information from $H$ heads:
\begin{equation}
\small
    \text{MultiHead}(Q, K, V) = \text{Concat}(\text{head}_1, ..., \text{head}_H)W_O
\end{equation}
where $\text{head}_i = \text{Attention}(QW_i^Q, KW_i^K, VW_i^V)$ and $W_O$ is the output projection. Finally, we obtain:
\begin{equation}
    h_{\text{det}}^{(l+1)} = \text{SMSA}(h_{\text{det}}^{(l)}, h_{\text{text}}^{(l)}, z_{\text{gen}}^{(l)}).
\end{equation}
This layer-wise interaction enables the detector to progressively perceive characteristics of the generative distribution. Finally, the detection head (a shallow MLP) outputs the predicted label $\hat{D} = F_{\text{det}}(h_{\text{det}}^{(L)})$, and the text decoding head generates the explanation $\hat{E}$, where $L$ denotes the total number of layers.

For the \textit{generation task}, samples contain real image $\mathcal{I}$ and text description $\mathcal{P}$. The generation module is based on Flow Matching pre-training. During training, ground truth images undergo a forward noising process to obtain noised latents:
\begin{equation}
    x_t = (1-t)x_0 + t\epsilon, \quad \epsilon \sim \mathcal{N}(0, I),
\end{equation}
where $t \in [0,1]$ is the flow time and $x_0$ represents the clean latent. We inject discriminative textual features extracted by the detector as conditions into the generation process to enhance the rationality of generated images.

\begin{figure}[t]
  \centering
  \includegraphics[width=\linewidth]{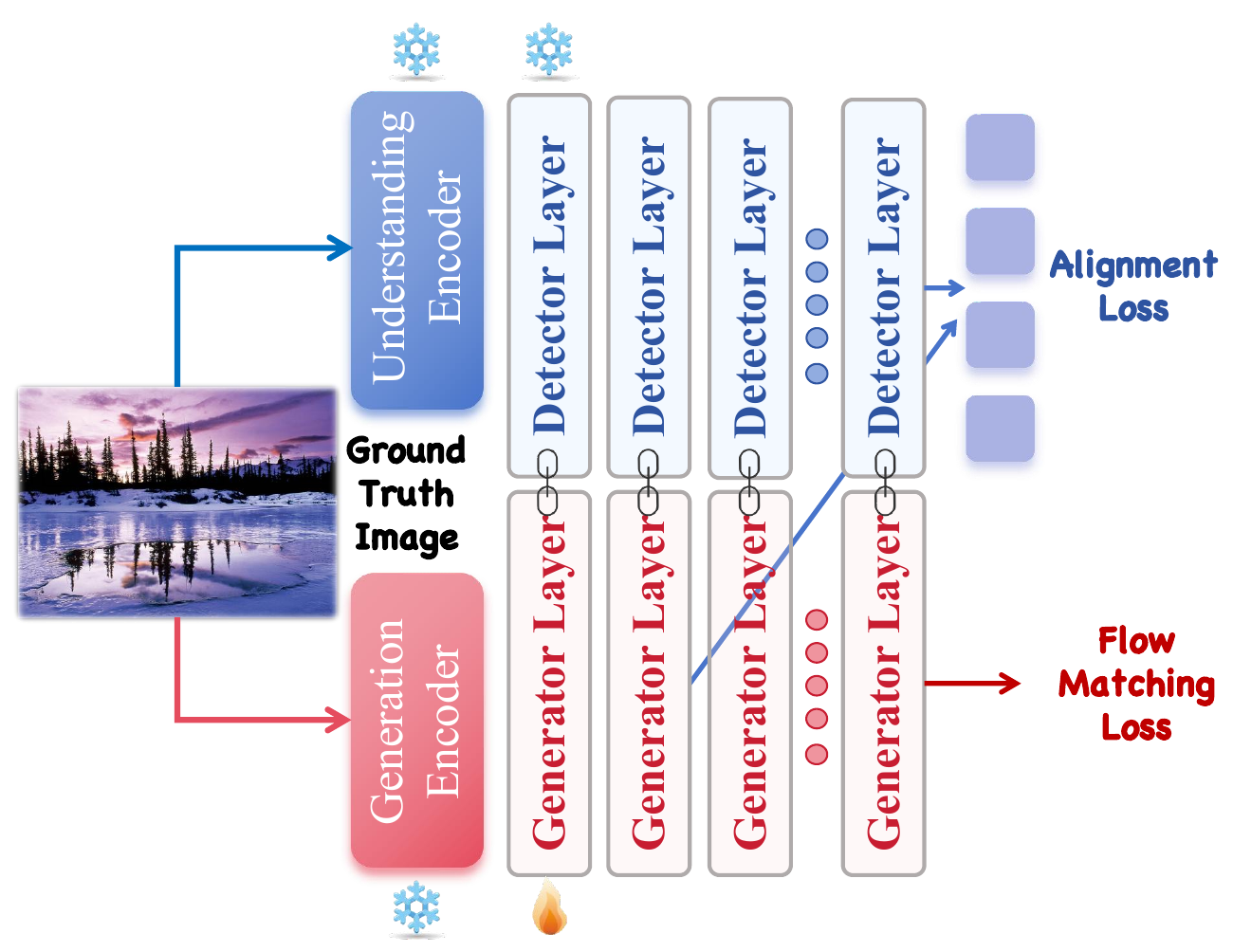}
  \vspace{-7mm}
  \caption{\textbf{Detector-Informed Generative Alignment (DIGA) pipeline.}
The generator (\textcolor{red}{\faFire}) learns from the frozen detector (\textcolor{blue}{\faSnowflake}) via feature alignment and flow matching.
This detector-informed alignment injects forensic knowledge into the generator, enabling authenticity-aware synthesis while preserving generative fidelity.
}
   \label{fig:pipeline2}
   \vspace{-5mm}
\end{figure}

\noindent\textbf{Loss Function Design.}
We design joint optimization objectives for different tasks. The detection classification loss uses binary cross-entropy:
\begin{equation}
\small
    \mathcal{L}_{\text{det}} = -\frac{1}{N}\sum_{i=1}^N \left[ y_i \log(\hat{D}_i) + (1-y_i)\log(1-\hat{D}_i) \right],
\end{equation}
where $y_i \in \{0,1\}$ is the ground truth label and $\hat{D}_i$ is the predicted probability. The explanation generation loss employs autoregressive language modeling:
\begin{equation}
    \mathcal{L}_{\text{exp}} = -\sum_{t=1}^T \log p_\theta(a_t | a_{<t}, h_{\text{det}}^L, h_{\text{text}}^L),
\end{equation}
where $a_t$ represents the $t$-th token in the explanation sequence. Full-parameter fine-tuning optimizes model parameters $\theta$ to enhance reasoning capability on synthetic data while maintaining instruction-following ability. The generation loss uses Flow Matching objective:
\begin{equation}
    \mathcal{L}_{\text{fm}} = \mathbb{E}_{t, x_0, x_t} \left[ \| v_\theta(x_t, t, c) - (x_0 - x_t) \|^2 \right],
\end{equation}
where $v_\theta(x_t, t, c)$ is the predicted velocity field conditioned on text $c$, and $(x_0 - x_t)$ is the target velocity. The total loss combines all three components:
\begin{equation}
    \mathcal{L} = \lambda_{\text{det}}\mathcal{L}_{\text{det}} + \lambda_{\text{exp}}  \mathcal{L}_{\text{exp}} + \lambda_{\text{fm}} \mathcal{L}_{\text{fm}},
\end{equation}
where $\lambda_{\text{det}}$, $\lambda_{\text{exp}}$, and $\lambda_{\text{fm}}$ are hyperparameters balancing the contribution of each task. Through multi-task joint optimization, we achieve collaborative improvement of both generation and detection capabilities.

\subsection{Detector-Informed Generative Alignment}
\label{sec:diga}

\noindent\textbf{Design Insights.} Current generative and detection research exists in a decoupled manner. Methods like LEGION \cite{kang2025legion} refine generation using detector signals at test time, but this is essentially a staged, high-latency post-processing enhancement that fails to fundamentally improve the generator's intrinsic authenticity. In other words, the generator remains unaware during training of which artifacts or features make it easily detectable. This decoupling creates a critical vulnerability: generators produce images with systematic biases that detectors can exploit, yet receive no feedback to correct these weaknesses.

Inspired by REPA \cite{yu2024representation}, which demonstrates that pretrained vision encoders such as DINOv2~\cite{oquab2023dinov2} can accelerate training through feature alignment, we leverage a more valuable tool: our specialized detector $f_D$ from GDUF. Unlike general encoders targeting semantic consistency, $f_D$ is trained to capture subtle forgery traces in AI-generated images, including frequency anomalies, texture inconsistencies, and imperceptible artifacts that distinguish synthetic from real content. Our insight focuses on proactively injecting the detector's forensic knowledge into the generator during training. Our Detector-Informed Generative Alignment (DIGA) forces the generator's features to align with perfectly real images from the detector's perspective, guiding it away from easily detectable feature subspaces. This creates a closed-loop training paradigm where the generator continuously learns what constitutes ``undetectable" characteristics.

\noindent\textbf{Architecture and Training.}
We reuse the frozen detector $f_D$ as an authenticity teacher to train generator $g_\theta$. For any ground-truth image $x_{\text{GT}}$, we perform parallel feature extraction: (1) We extract patch-wise features $z_D \in \mathbb{R}^{N \times C_D}$ from $f_D$'s final Transformer block, which encode the detector's high-level perception of authenticity. (2) We extract intermediate features $z_G = g_\theta^{(l)}(z_t, t) \in \mathbb{R}^{N \times C_G}$ from layer $l$ of the generator, given noisy input $z_t$ at timestep $t$. These mid-level features possess sufficient semantic structure while remaining malleable for alignment guidance. A lightweight trainable projection $h_\phi$ bridges the dimensional gap: $\hat{z}_G = h_\phi(z_G)$. Our DIGA loss distills generator features to match detector representations via cosine similarity as follows:
\begin{equation}
\small
\mathcal{L}_{\text{DIGA}} = \mathbb{E}_{x_{\text{GT}}, z_t, t} \left[ 1 - \frac{h_\phi(g(z_t, t)) \cdot f_D(x_{\text{GT}})}{\|h_\phi(g(z_t, t))\| \|f_D(x_{\text{GT}})\|} \right],
\end{equation}
combined with Flow Matching loss:
\begin{equation}
\mathcal{L}_{\text{total}} = \mathcal{L}_{\text{flow}} + \lambda \mathcal{L}_{\text{DIGA}}.
\end{equation}

Through this joint optimization, our framework effectively internalizes detection awareness into the generator's learned representations, simultaneously enhancing both the visual authenticity of generated images and their robustness against forensic analysis, while maintaining the original generative capabilities.
\section{Experiments}

This section systematically evaluates UniGenDet's performance from five aspects: experimental setup verification, synthetic image detection capability, cross-dataset generalization on the FakeClue benchmark, generation quality assessment, and component-wise ablation studies. Both quantitative metrics and qualitative results demonstrate our framework's advantages in bridging generation and detection tasks.

\subsection{Experimental Setups}

\noindent\textbf{Datasets.}
To comprehensively evaluate our model's performance in both forgery detection and image generation tasks, we carefully selected specialized datasets for each domain. For forgery detection, we employ the FakeClue~\cite{wen2025spot} dataset for training, which contains diverse types of synthetic images with detailed annotations, enabling the model to effectively learn and understand authenticity cues. For generation tasks, following BAGEL~\cite{deng2025emerging}, we use the high-aesthetic subset of LAION~\cite{schuhmann2022laion}, comprising 80K images, to ensure high-quality training data. For evaluation, we utilize three challenging benchmarks: 
FakeClue test set,
DMimage~\cite{corvi2023detection}
, and ARForensics~\cite{zhang2025d3qe}.
To assess generation quality, we conduct experiments to evaluate the Frechet Inception Distance (FID)~\cite{heusel2017gans}, a metric that quantifies the similarity between the distributions of generated and real images in a deep feature space. We then adopt the GenEVAL~\cite{ghosh2023geneval} evaluation set, which includes human-preferred metrics for evaluating both image fidelity and text-image alignment.

\noindent\textbf{Baselines.}
For forgery detection, we compare against both pre-trained large multimodal models and specialized detection methods. Pre-trained LMMs include: DeepSeek-VL2-small~\cite{wu2024deepseekvl2mixtureofexpertsvisionlanguagemodels}, DeepSeek-VL2~\cite{wu2024deepseekvl2mixtureofexpertsvisionlanguagemodels}, InternVL2-8B~\cite{chen2024internvl}, InternVL2-40B~\cite{chen2024internvl}, Qwen2-VL-7B~\cite{wang2024qwen2}, Qwen2-VL-72B~\cite{wang2024qwen2}, GPT-4o (2024-08-06)~\cite{openai2025chatgpt4o}, and BAGEL~\cite{deng2025emerging}. Specialized detectors include: CNNSpot~\cite{wang2020cnn}, FreqNet~\cite{tan2024frequency}, Fatformer~\cite{liu2024forgery}, UnivFD~\cite{ojha2023towards}, AIDE~\cite{yan2024sanity}, NPR~\cite{tan2024rethinking}, $\bf{D^3}$QE~\cite{zhang2025d3qe}, SIDA~\cite{huang2024sida}, AntifakePrompt~\cite{chang2023antifakeprompt} and FakeVLM~\cite{wen2025spot}. All baselines were evaluated using official source code and recommended parameter configurations to ensure fair comparison.

\noindent\textbf{Implementation Details.}
Our framework is built upon the pretrained generative-understanding unified model BAGEL. The Understanding Encoder uses SigLIP2~\cite{tschannen2025siglip}, while the Generation Encoder employs the pre-trained VAE from FLUX~\cite{flux}. The generation and detection branches are initialized with separate Qwen2.5 LLM~\cite{qwen2.5-vl} weights. All experiments were conducted on 8 NVIDIA A100 GPUs. During the first training stage, we maintain a 1:1 ratio between generation and understanding data within each batch, with a total batch token count of $32768 \times 8$. Training utilizes the AdamW optimizer with a learning rate of $1\times10^{-4}$ and a weight decay of $1\times10^{-2}$ for 2 epochs. The hyperparameters for the different loss weights $\lambda_{\text{det}}$, $\lambda_{\text{exp}}$, and $\lambda_{\text{fm}}$ are all set to 1 to balance detection, explanation, and generation objectives.

\subsection{Evaluation on Synthetic Image Detection}

\begin{table}[!t]
\caption{
\textbf{Comparison of both detection and artifact explanation performance on the \textbf{FakeClue} dataset. }
$*$ denotes methods trained on FakeClue. 
$\dagger$ indicates results re-evaluated using the official checkpoint under our unified protocol.
}
\vspace{-3mm}
\begin{adjustbox}{width=\linewidth,center}
\setlength{\tabcolsep}{2pt}
\begin{tabular}{rccccc}
\hline
\addlinespace[1pt]
  \multirow{2}{*}{Type} &  \multirow{2}{*}{Method}  & \multicolumn{4}{c}{FakeClue}  \\
   \cmidrule(r){3-6}
   && Acc $\uparrow$ & F1 $\uparrow$ & ROUGE\_L $\uparrow$& CSS $\uparrow$ \\
   \hline
   \addlinespace[1.5pt]
   \multirow{8}{*}{\textbf{I}} 
   & Deepseek-VL2-small \cite{wu2024deepseekvl2mixtureofexpertsvisionlanguagemodels} & $40.4$ & $54.2$ & $17.1$ & $50.4$ \\
   & Deepseek-VL2 \cite{wu2024deepseekvl2mixtureofexpertsvisionlanguagemodels} & $47.5$ & $54.1$ & $17.2$ & $50.5$ \\
   & InternVL2-8B \cite{chen2024internvl}  & $50.6$ & $49.0$ & $18.0$ & $58.1$ \\
   & InternVL2-40B \cite{chen2024internvl}  & $50.7$ & $46.3$ & $17.6$ & $55.2$ \\
   & Qwen2-VL-7B \cite{wang2024qwen2} & $45.7$ & $59.2$ & $26.6$ & $56.5$ \\
   & Qwen2-VL-72B \cite{wang2024qwen2} & $57.8$ & $56.5$ & $17.5$ & $54.4$ \\
   & GPT-4o (2024-08-06) \cite{openai2025chatgpt4o} & $47.4$ & $42.0$ & $13.4$ & $40.7$ \\
   & BAGEL \cite{deng2025emerging} & $40.5$ & $34.1$ & $23.9$ & $46.2$ \\
    \midrule
   \multirow{8}{*}{\textbf{II}} 
   & CNNSpot \cite{wang2020cnn} & $43.1$ & $9.8$ & - & - \\
   & FreqNet \cite{tan2024frequency} & $48.7$ & $39.3$ & - & - \\
   & Fatformer \cite{liu2024forgery} & $54.5$ & $45.1$ & - & - \\
   & UnivFD \cite{ojha2023towards} & $63.1$ & $46.8$ & - & - \\
   & AIDE$^{*}$ \cite{yan2024sanity} & $85.9$ & $94.5$ & - & - \\
   & NPR$^{*}$ \cite{tan2024rethinking} & $90.2$ & $91.6$ & - & - \\
   & FakeVLM$^{*,\dagger}$~\cite{wen2025spot} & $\mathbf{98.6}$ & $\mathbf{98.1}$ & ${32.2}$ & ${59.5}$ \\
     & Ours$^{*}$ &  $98.0$ & $97.7$ & $\mathbf{56.3}$ & $\mathbf{79.8}$ \\
\hline
\end{tabular}
\end{adjustbox}
\label{tab:maintable}
\vspace{-3mm}
\end{table}

\begin{table}[t]
\caption{\textbf{Comparison with other detection methods on the DMimage \cite{corvi2023detection} dataset,} using the original weights for each method.}
\vspace{-3mm}
\renewcommand{\arraystretch}{1.05}
\setlength{\tabcolsep}{5pt}
\centering
\resizebox{\linewidth}{!}{ 
\begin{tabular}{ccccccc}
\toprule
\multirow{2}{*}{Method} & \multicolumn{2}{c}{Real} & \multicolumn{2}{c}{Fake} & \multicolumn{2}{c}{Overall} \\ 
\cmidrule(r){2-3} \cmidrule(r){4-5} \cmidrule(r){6-7}
 & Acc & F1 & Acc & F1 & Acc & F1 \\ \midrule
CNNSpot \cite{wang2020cnn}      &  87.8   & 88.4   &  28.4  & 44.2  &  40.6  &  43.3   \\
Gram-Net \cite{gramnet}     & 62.8    & 54.1   & 78.8  & 88.1  &  67.4  &  79.4   \\
Fusing \cite{ju2022fusing}         &  87.7   & 86.1    & 15.5    & 27.2  &  40.4  &  36.5   \\
LNP \cite{LNP}            & 63.1    & 67.4   & 56.9   & 72.5  &  58.2  &  68.3   \\
UnivFD \cite{ojha2023towards}       & 89.4    & 88.3   & 44.9   & 61.2  &  53.9  &  60.7   \\
AntifakePrompt \cite{chang2023antifakeprompt} &  91.3    & 92.5    & 89.3    & 91.2  &  90.6  &  91.2   \\
SIDA \cite{huang2024sida} &  92.9 & 93.1 & 90.7 & 91.0 & 91.8 & 92.4 \\
FakeVLM  \cite{wen2025spot}        & 98.2    & 99.1   & 89.7   &  94.6  &  94.0  &  94.3 \\ \midrule
UniGenDet       & \textbf{99.0}    & \textbf{99.5}   & \textbf{97.2}   &  \textbf{98.6}  &  \textbf{98.6}  &  \textbf{99.1} \\ \bottomrule
\end{tabular}
}
\label{table:dmimage}
\vspace{-3mm}
\end{table}

\noindent\textbf{Evaluation on the FakeClue Dataset.} We begin with a systematic evaluation of the proposed model’s capability for synthetic image detection. Table~\ref{tab:maintable} compares UniGenDet with current mainstream approaches, including large language models (Category I) and specialized detection models (Category II). Performance is assessed along four dimensions: Accuracy (Acc), F1 score, ROUGE\_L, and CSS—where Acc and F1 reflect detection discriminability, ROUGE\_L measures the match between generated explanations and reference answers, and CSS evaluates semantic consistency. On the FakeClue dataset, UniGenDet achieves outstanding results, with a detection accuracy of 98.0\% and an F1 score of 97.7\%. These figures represent improvements of 40.2\% in Acc and 41.2\% in F1 over the strongest open-source model, Qwen2-VL-72B. Compared to expert models also trained on FakeClue, our method surpasses NPR by 7.8\% in Acc and AIDE by 12.1\% in Acc. As FakeVLM does not include official implementations for ROUGE-L and CSS, we re-evaluated it using the provided checkpoint following our unified evaluation protocol. Although it shows marginally lower accuracy, it significantly outperforms in text-related metrics, highlighting its superior interpretability and cross-modal reasoning ability.

\noindent\textbf{Cross-Dataset Evaluation.} To examine the generalization of various models, we conduct cross-dataset testing on the DMImage dataset. As shown in Table~\ref{table:dmimage}, UniGenDet achieves top performance on both real and fake image detection, with an overall accuracy of 98.6\% and an F1 score of 99.1\%, clearly outperforming the previous best method SIDA (+6.8\% Acc, +6.7\% F1). It is worth emphasizing that our approach does not rely on any external classifiers or expert models; while retaining natural language explanation capabilities, it exceeds the performance of specialized detection models, demonstrating the advantage of the unified architecture. 

As shown in Table~\ref{tab:AR}, to assess generalization against newer model architectures, we evaluated our method on the ARForensics dataset, which features images from state-of-the-art visual autoregressive generators. In this challenging zero-shot setting, UniGenDet achieves a mean accuracy of 98.1\%. This result significantly outperforms prior specialized detectors (e.g., $\bf{D^3}$QE at 82.1\%) and surpasses the strong FakeVLM baseline (97.1\%). The superior performance confirms UniGenDet's robust generalization capabilities and its resilience to rapidly advancing generative paradigms.

\definecolor{mygray}{gray}{0.5} 

\begin{table}[t]
  \caption{\textbf{Generalization performance on the ARForensics dataset.} 
  Acc. denotes accuracy. Roman numerals indicate training/evaluation setup: \textbf{I} – trained on the LlamaGen dataset; \textbf{II} – zero-shot evaluation. (OM2 refers to the Open-MAGVIT2 model.)}
  \vspace{-3mm}
  \centering
  \setlength{\tabcolsep}{2pt}
  \renewcommand{\arraystretch}{1.1}
  \resizebox{\linewidth}{!}{
  \begin{tabular}{l l c c c c c c c c}
  \bottomrule \hline
  \multirow{2}*{\centering Type} & \multirow{2}*{\centering Method} & {LlamaGen} & {VAR} & {Infinity} & {Janus-Pro} & {RAR} & {Switti} & {OM2} & {Mean}\\
  & & Acc. & Acc. & Acc. & Acc. & Acc. & Acc. & Acc. & Acc.\\ 
  \bottomrule \hline
\multirow{7}{*}{\textbf{I}} 
& CNNSpot~\cite{wang2020cnn}  & 99.9 & 50.3 & 50.9 & 95.7 & 50.8 & 56.6 & 50.1 & 64.9 \\
& FreDect~\cite{frank2020leveraging} & 99.8 & 52.9 & 50.2 & 88.9 & 52.5 & 50.0 & 57.1 & 64.5 \\
& Gram-Net~\cite{gramnet} & 99.6 & 55.0 & 52.4 & 74.5 & 50.0 & 57.7 & 50.1 & 62.8 \\
& LNP~\cite{liu2022detecting} & 99.5 & 49.6 & 49.8 & 99.5 & 49.7 & 70.3 & 49.6 & 66.9 \\
& UnivFD~\cite{ojha2023towards} & 89.9 & 80.5 & 71.7 & 84.3 & 88.3 & 76.0 & 66.2 & 79.6 \\
& NPR~\cite{tan2024rethinking} & \textbf{100.0} & 56.9 & 88.5 & 93.7 & 52.3 & 52.0 & 63.0 & 72.3 \\
& D$^3$QE~\cite{zhang2025d3qe} & 97.2 & 85.3 & 62.9 & 92.3 & 91.7 & 75.3 & 70.1 & 82.1 \\
\hline
\multirow{2}{*}{\textbf{II}}
& FakeVLM~\cite{wen2025spot} & 98.1 & 97.7 & 99.4 & \textbf{99.9} & \textbf{99.9} & 94.6 & 90.4 & 97.1 \\
& \textbf{UniGenDet} & 89.4 & \textbf{99.7} & \textbf{99.9} & 99.7 & 99.5 & \textbf{98.8} & \textbf{99.5} & \textbf{98.1} \\
\bottomrule
  \end{tabular}
  }
  \label{tab:AR}
  \vspace{-3mm}
\end{table}

\subsection{Generation Evaluation}

To comprehensively assess the model’s generation performance, we designed two sets of experiments evaluating image quality and text–image alignment, respectively.

\noindent\textbf{Image Quality Evaluation.} We quantitatively analyze the visual quality of generated images using the widely-adopted Fréchet Inception Distance (FID), which measures the perceptual similarity between the distributions of synthetic and real images in a deep feature space. For this evaluation, we randomly selected 5,000 text prompts from the LAION dataset, ensuring they were disjoint from the training set to guarantee a fair assessment. We then generated corresponding images to compute the FID score relative to the real image distribution. As shown in Table~\ref{tab:fid_results}, UniGenDet achieves the best FID score of 17.5, significantly outperforming the original BAGEL (22.9) and the intermediate BAGEL+GDUF (19.4). This improvement confirms that injecting discriminative knowledge about real and fake images helps the generator produce more realistic and visually plausible samples. The lower FID indicates that our model not only captures the global distribution of natural images more accurately but also reduces artifacts and inconsistencies commonly observed in synthetic outputs, thereby enhancing overall generation fidelity.

\begin{table}[t]
\caption{\textbf{Image quality comparison using Fréchet Inception Distance (FID).} 
We evaluate the realism and fidelity of generated images across different model variants. }
\vspace{-3mm}
\centering
\small
\setlength{\tabcolsep}{6pt}
\begin{tabular}{lccc}
\hline
\textbf{Model} & \textbf{BAGEL} & \textbf{BAGEL+GDUF} & \textbf{UniGenDet} \\
\hline
\textbf{FID} $\downarrow$ & 22.9 & 19.4 & \textbf{17.5} \\
\hline
\end{tabular}
\vspace{-2mm}
\label{tab:fid_results}
\end{table}

\begin{table}[t]
    \centering
    \setlength{\tabcolsep}{4pt}
    \caption{\textbf{Evaluation of text-to-image generation on the GenEval benchmark.} 
‘Gen. Only’ denotes pure generation models, while ‘Unified’ refers to models with both understanding and generation capabilities. 
$\dagger$ indicates methods employing an LLM-based rewriter. 
SO, TO, CL, ATTR, and POS correspond to {Single Object}, {Two Object}, {Colors}, {Color Attribute}, {Position}, and {Counting}, respectively.
}
\vspace{-3mm}
\resizebox{\linewidth}{!}{
    \begin{tabular}{clcccccc}
        \toprule
        \textbf{Type} & \textbf{Model}  & \textbf{SO} & \textbf{TO} & \textbf{CT} & \textbf{CL} & \textbf{POS} & \textbf{ATTR}  \\
        \midrule
        \multirow{7}{*}{\rotatebox{90}{\textit{Gen. Only}}}
        & PixArt-$\alpha$~\cite{chen2024pixart} &  0.98 & 0.50 & 0.44 & 0.80 & 0.08 & 0.07  \\
        & SDv$2.1$~\cite{rombach2022high} & 0.98 & 0.51 & 0.44 & 0.85 & 0.07 & 0.17  \\
        & DALL-E $2$~\cite{dalle2}  & 0.94 & 0.66 & 0.49 & 0.77 & 0.10 & 0.19 \\
        & SDXL~\cite{sdxl} &  0.98 & 0.74 & 0.39 & 0.85 & 0.15 & 0.23 \\
        & DALL-E $3$~\cite{dalle3} & 0.96 & 0.87 & 0.47 & 0.83 & 0.43 & 0.45  \\
        & SD3-Medium~\cite{SD3} & \textbf{0.99} & 0.94 & 0.72 & 0.89 & 0.33 & 0.60  \\
        & FLUX.1-dev$^{\dagger}$~\cite{flux} & 0.98 & 0.93 & 0.75 & 0.93 & 0.68 & 0.65  \\
        \midrule
        \multirow{7}{*}{\rotatebox{90}{\textit{Unified}}}
        & TokenFlow-XL~\cite{qu2024tokenflow} &  0.95 & 0.60 & 0.41 & 0.81 & 0.16 & 0.24  \\
        & Janus~\cite{janus2024} & 0.97 & 0.68 & 0.30 & 0.84 & 0.46 & 0.42  \\
        & Emu$3$-Gen$^{\dagger}$\cite{emu3} & \textbf{0.99} & 0.81 & 0.42 & 0.80 & 0.49 & 0.45  \\
        & Show-o~\cite{show-o} &  0.98 & 0.80 & 0.66 & 0.84 & 0.31 & 0.50  \\
        & Janus-Pro-7B~\cite{januspro2025} &  0.99 & 0.89 & 0.59 & 0.90 &\textbf{ 0.79} & 0.66  \\
    &   {BAGEL}$^{\dagger}$~\cite{deng2025emerging} & 0.98 &  \textbf{0.97}  &  \textbf{0.86} &  0.93 &  0.74 &  \textbf{0.77}   \\
    &  \textbf{UniGenDet}$^{\dagger}$ &  \textbf{0.99} &  0.95  &  0.80 &  \textbf{0.94} &  0.74 &  0.75  \\
    \bottomrule
    \end{tabular}}
    \vspace{-2mm}
    \label{tab:geneval}
\end{table}

\begin{figure*}[t]
  \centering
  \includegraphics[width=\linewidth]{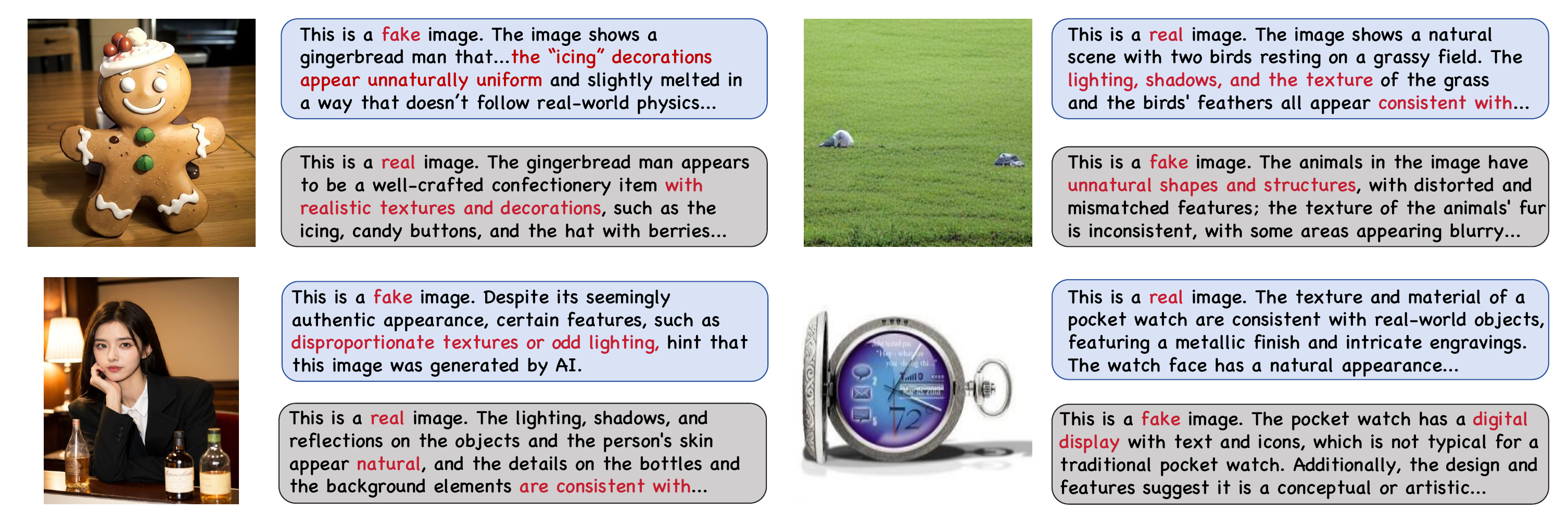}
  \vspace{-5mm}
  \caption{\textbf{Comparison of detection results.} For each sample (left: generated, right: real), the UniGenDet (top) outperforms the pretrained BAGEL (bottom), providing more accurate detection and superior explanation of artifacts in fake images and features in real ones.}
   \label{fig:comp}
   \vspace{-0.5mm}
\end{figure*}

\begin{figure}[t]
  \centering
  \includegraphics[width=\linewidth]{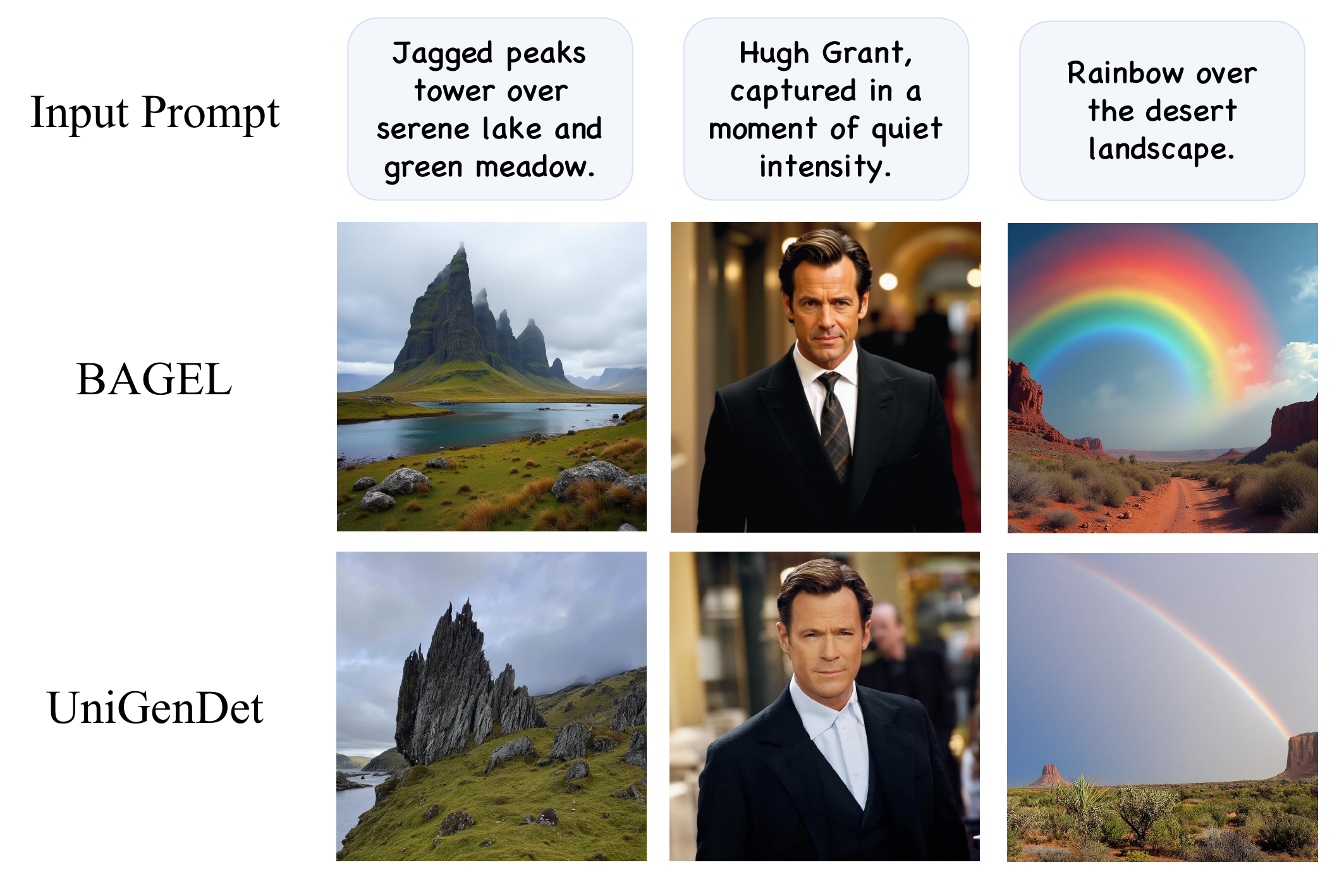}
  \vspace{-4mm}
  \caption{BAGEL (middle) vs UniGenDet (bottom) generation comparison. UniGenDet produces more natural landscapes with coherent lighting, validating detection-guided optimization.}
   \label{fig:vis}
   \vspace{-4mm}
\end{figure}
\noindent\textbf{Text-Image Alignment Assessment.} We further evaluate fine-grained text–image alignment and reasoning ability on the GenEval~\cite{ghosh2023geneval} benchmark in Table ~\ref{tab:geneval}. GenEval contains over 500 carefully designed prompts and assesses text-to-image generation across six dimensions. As summarized in Table~\ref{tab:geneval}, UniGenDet achieves an average score of 0.86 among unified models, performing comparably to the original BAGEL (0.87). It obtains the best score in SO (0.99) and CL (0.94), demonstrating strong basic object generation and multi-object reasoning capabilities. Although slightly lower than BAGEL in CT and ATTR, this minor gap reflects an acceptable trade-off in multi-task learning—maintaining powerful forgery detection without substantial degradation in generation quality. Since authenticity discrimination primarily emphasizes distributional realism rather than text–image correspondence, improvements in GenEval alignment metrics are not necessarily expected. Compared to specialized generative models, UniGenDet remains highly competitive, outperforming most dedicated generators in TO (0.95) and CL (0.94). These results validate that our unified framework effectively balances and enhances both generation and detection, achieving synergistic progress in dual-task learning.

\subsection{Ablation Study}
\begin{table}[t]
\caption{Comparison of different configurations. Higher is better.}
\vspace{-3mm}
\centering
\small
\setlength{\tabcolsep}{6pt}
\begin{tabular}{lcccc}
\hline
\textbf{Model} & \textbf{Acc} $\uparrow$ & \textbf{F1} $\uparrow$ & \textbf{Rouge-L} $\uparrow$ & \textbf{CSS} $\uparrow$ \\
\hline
w/o GDUF & 40.5 & 34.1 & 23.9 & 46.2 \\
w/o SMSA & 95.0 & 94.6 & 50.9 & 77.7 \\
 UniGenDet & \textbf{98.0} & \textbf{97.7} & \textbf{56.3} & \textbf{79.8} \\
\hline
\end{tabular}
\vspace{-5mm}
\label{tab:ablation}
\end{table}
To validate the contribution of each component, we conducted systematic ablation experiments under identical training settings. 

Table~\ref{tab:ablation} presents the results on the FakeClue test set. First, when only the first-stage unified fine-tuning of generation and understanding is performed (without GDUF optimization), the model achieves limited performance. Detection accuracy improves to 98.0\% (+57.5) and F1 to 97.7\% (+63.6) in the complete model, demonstrating that optimizing the generator based on the trained detector in the second stage plays a crucial role in improving authenticity discrimination. Second, removing the Symbiotic Multi-modal Self-Attention (SMSA) module leads to a notable drop in performance: accuracy decreases to 95.0\% (–3.0\%), F1 to 94.6\% (–3.1\%), and ROUGE-L by 5.4 points. 
This confirms the importance of cross-modal interaction established in the first stage—SMSA enables the detector to leverage the generator’s deep understanding of image distribution, thereby improving both detection precision and interpretability.


In terms of generation quality, Table~\ref{tab:fid_results} provides strong evidence: the full two-stage model achieves the best FID score of 17.5, outperforming both the first-stage baseline (22.9) and the intermediate BAGEL+GDUF version (19.4). 
These results verify the synergistic effect of two-stage training—while the first stage establishes a unified foundation, the second stage enhances generation quality through authenticity-guided feedback, leading to more natural and artifact-free outputs. 
Overall, the ablation results demonstrate that the two-stage framework and SMSA module together form an effective bidirectional enhancement mechanism.

\subsection{Qualitative Results}

As shown in Figure~\ref{fig:comp}, we present a comparative visualization of detection results and explanatory rationales in image authenticity identification. Each sample group follows a left-fake, right-real arrangement, with UniGenDet’s detection and explanation outputs displayed above and the fine-tuned BAGEL model’s results below. Experiments demonstrate that UniGenDet performs more robustly in discriminating between real and fake images—not only delivering more accurate detection judgments but also providing clearer and more reasonable explanations for artifacts in generated images (e.g., implausible texture proportions, irregular lighting) and natural characteristics in real images. For instance, in the top-left generated gingerbread figure image, despite its realistic appearance, UniGenDet accurately identifies disproportioned textures and lighting inconsistencies. In contrast, the BAGEL model not only shows limited sensitivity to synthetic traces but also misinterprets legitimate elements in real images (such as the digital display on a pocket watch) as anomalies, reflecting its insufficient comprehension.

Figure~\ref{fig:vis} compares the inference-stage outputs of BAGEL and UniGenDet under identical input prompts and parameter settings. It can be observed that images generated by BAGEL exhibit several implausible artifacts—over-smoothed grassland, physically inconsistent lake reflections in landscape examples, and unnatural surface gloss in character portraits. In comparison, samples generated by UniGenDet, fine-tuned under the guidance of the authenticity discrimination module, are visually closer to real images, demonstrating a marked improvement in both structural coherence and naturalness, which validates the effectiveness of the detection module in enhancing generation quality.

\section{Conclusion}
This paper has introduced UniGenDet, a unified framework that bridges image generation and authenticity detection through co-evolutionary training. Our approach demonstrates that generation and detection can mutually enhance each other: detection feedback refines generation realism, while generation improves artifact identification and interpretability. Extensive experiments demonstrate state-of-the-art performance in both tasks, significantly outperforming isolated models. This work provides an effective solution to the escalating arms race between generative AI and generated content authentication. 

\newpage
\section*{Acknowledgments}

This work was supported in part by the National Natural Science Foundation of China under Grant 62441616, Grant 62336004, Grant 62125603, Grant 62306031, Grant 62406172, Grant 62506198, in part by the China Postdoctoral Science Foundation under Grant 2024M761674 and Grant 2023M741964, and in part by the Postdoctoral Fellowship Program of CPSF No. GZC20240841.

{\centering
        \Large
        \vspace{1.5em}Supplementary Material \\
        \vspace{0.5em}}
        
\renewcommand\thesection{\Alph{section}} 
\renewcommand\thetable{\Alph{table}}
\renewcommand\thefigure{\Alph{figure}}
\setcounter{section}{0}
\setcounter{figure}{0}
\setcounter{table}{0}


\begin{figure*}[h]
  \centering
  \includegraphics[width=\linewidth]{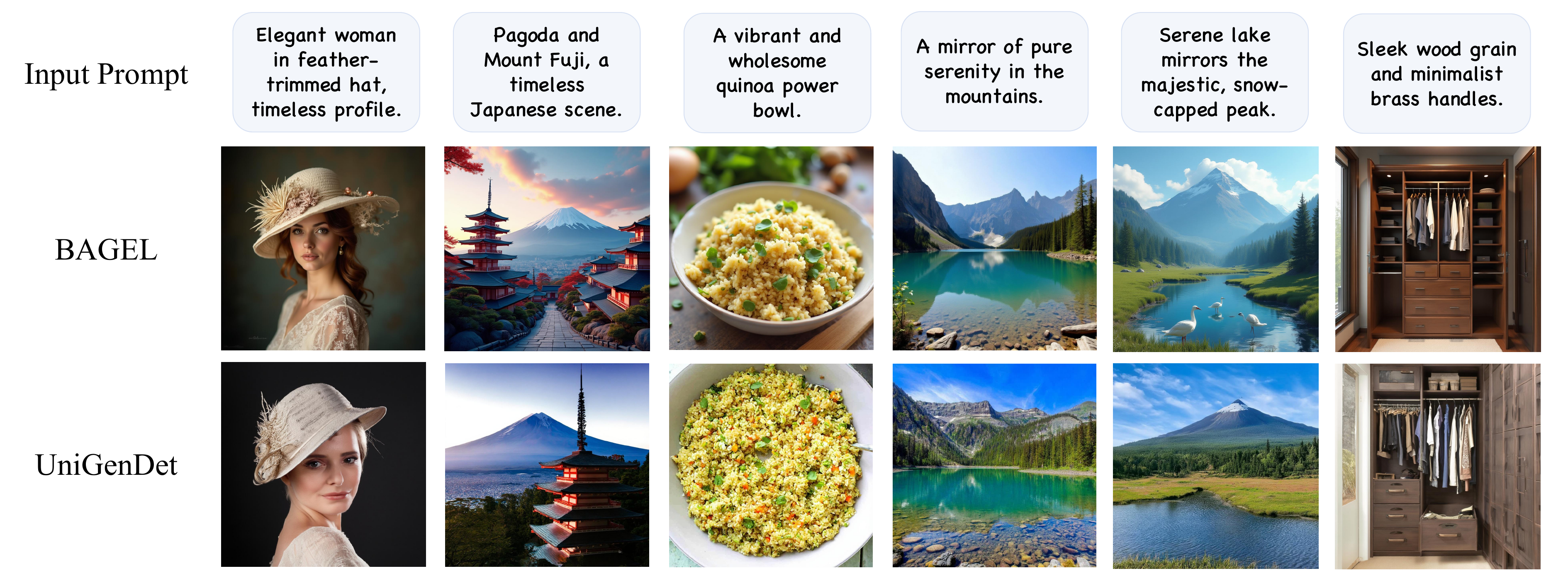}
  \vspace{-5mm}
  \caption{\textbf{Comparison of generation result}. BAGEL (middle) vs UniGenDet (bottom) generation comparison. UniGenDet produces more natural landscapes with coherent lighting, validating detection-guided optimization.}
   \label{fig:comp_supp}
   \vspace{-3mm}
\end{figure*}

\section{Implementation Details}
\label{sec:impl}

In this section, we provide comprehensive details regarding the training pipeline, model configurations, and inference settings to facilitate reproducibility.

\subsection{Training Setup and Hyperparameters}

The training process of   UniGenDet is divided into two phases: Generation-Detection Unified Fine-tuning (GDUF) and Detector-Informed Generative Alignment (DIGA).

\textbf{GDUF Stage.} The complete training pipeline of the GDUF stage requires approximately 12 hours on 8 NVIDIA A100 GPUs. The model is trained for about 1,000 optimization steps with a total batch token count of $16384 \times 8$.

\textbf{DIGA Stage.} The subsequent DIGA training phase completes in approximately 6 hours over 500 steps s on 8 NVIDIA A100 GPUs. During this stage, we specifically align the features from the 8th layer of the generator with the final layer features of the detection module. The alignment loss weight is set to $\lambda = 0.5$. To ensure training stability and effectiveness, we maintain a balanced ratio of positive (real) and negative (fake) samples throughout the alignment process.

\textbf{Data Preprocessing.} Following the configuration of the base model BAGEL~\cite{deng2025emerging}, we apply distinct preprocessing strategies for different tasks:
\begin{itemize}
    \item \textbf{Detection Inputs:} Processed with a patch extraction strategy of [min=224, max=980, stride=14] using the SigLIP encoder.
    \item \textbf{Generation Inputs:} Processed with a patch extraction strategy of [min=512, max=1024, stride=16] for the VAE encoder.
\end{itemize}

\subsection{Model Architectures}

Our framework is built upon a powerful Large Language Model (LLM) and a high-resolution vision encoder. Detailed architectural configurations are listed in Table~\ref{tab:model_config}. The visual encoder is based on a SigLIP~\cite{tschannen2025siglip} architecture, while the LLM backbone is initialized from Qwen2.5~\cite{qwen2.5-vl} weights.

\begin{table}[h]
\centering
\caption{\textbf{Detailed Model Configurations.} Specification of the Vision Encoder and LLM Backbone used in UniGenDet.}
\vspace{-3mm}
\label{tab:model_config}
\setlength{\tabcolsep}{2.8pt}
\resizebox{\linewidth}{!}{
\begin{tabular}{c|cc}
\toprule
\textbf{Component} & \textbf{Parameter} & \textbf{Value} \\
\midrule
& Hidden Size & 3,584 \\
\multirow{3}{*}{\textbf{LLM Backbone}} & Intermediate Size & 18,944 \\
\multirow{3}{*}{(Qwen2.5)} & Number of Layers & 28 \\
& Attention Heads & 28 \\
& Key/Value Heads & 4 \\
& Vocab Size & 152,064 \\
\midrule
& Model Type & SigLIP Vision Model \\
\multirow{2}{*}{\textbf{Vision Encoder} } & Image Size & 980 \\
\multirow{3}{*}{(SigLIP)} & Patch Size & 14 \\
& Hidden Size & 1,152 \\
& Number of Layers & 27 \\
\bottomrule
\end{tabular}
}
\vspace{-5mm}
\end{table}

\subsection{Inference Settings}
For the text-to-image generation task, we employ a 50-step diffusion sampling process. For text generation (including detection explanations), we use nucleus sampling with a temperature of $0.7$, top-$p$ of 0.8, and top-$k$ of 20. A repetition penalty of $1.05$ is applied to prevent degenerative loops in the generated explanations.

\section{Attention Mask Mechanism}
\label{sec:attn_mask}

A core design of UniGenDet is the task-specific attention masking strategy that orchestrates the flow of information between text, visual understanding features, and generative latents.


\textbf{Generation Task.} 
In the text-to-image generation mode, the input sequence consists of text prompts followed by VAE latent noise. 
\begin{itemize}
    \item \textbf{Text Tokens:} We apply a \textit{causal mask} (attending only to preceding tokens) to preserve the autoregressive nature of language modeling.
    \item \textbf{Visual Latents:} We employ a \textit{bidirectional mask} within the image sequence, allowing image noise tokens to attend to all other image tokens and all preceding text tokens. This ensures the generated visual content aligns with the textual prompt.
\end{itemize}

\textbf{Detection Task.} 
The generated image detection task involves a more complex interaction among three components: the Generation Encoder (VAE), the Detection Encoder (ViT), and the Text Instructions.
\begin{itemize}
    \item \textbf{VAE Latents ($z_{gen}$):} These tokens attend globally to themselves to model the generative distribution.
    \item \textbf{ViT Features ($h_{det}$):} Through the Symbiotic Multi-modal Self-Attention (SMSA) module, these features attend globally to themselves and cross-attend to the VAE latents, enabling the detector to perceive generative artifacts.
    \item \textbf{Text Tokens:} The instruction and answer tokens use a \textit{causal mask} internally. Importantly, they can attend to \textbf{all} preceding visual tokens (both VAE and ViT features) to perform grounded reasoning and generate explanations.
\end{itemize}



\begin{table}[t]
\centering
\caption{Robustness Comparison under JPEG Compression.}
\label{tab:jpeg_robustness}
\vspace{-3mm}
\setlength{\tabcolsep}{2pt}
\resizebox{\linewidth}{!}{
\begin{tabular}{ccccccc}
\toprule
JPEG Quality & Method & Accuracy↑ & F1-Score↑ & ROUGE-L↑ & CSS↑ \\
\midrule
\multirow{2}{*}{\textit{N/A}} & FakeVLM & $\mathbf{98.6}$ & $\mathbf{98.1}$ & $32.2$ & $59.5$ \\
 & Ours & $98.0$ & $97.7$ & $\mathbf{56.3}$ & $\mathbf{79.8}$ \\
\midrule
\multirow{2}{*}{90} & FakeVLM & $86.8$ & $86.4$ & $29.5$ & $54.2$ \\
 & Ours & $\mathbf{95.1}$ & $\mathbf{94.7}$ & $\mathbf{54.0}$ & $\mathbf{76.7}$ \\
\midrule
\multirow{2}{*}{70} & FakeVLM & $88.4$ & $88.0$ & $30.2$ & $54.8$ \\
 & Ours & $\mathbf{91.4}$ & $\mathbf{90.8}$ & $\mathbf{52.1}$ & $\mathbf{75.1}$ \\
\midrule
\multirow{2}{*}{50} & FakeVLM & $80.4$ & $80.3$ & $29.3$ & $53.2$ \\
 & Ours & $\mathbf{91.3}$ & $\mathbf{90.7}$ & $\mathbf{51.7}$ & $\mathbf{74.4}$ \\
\bottomrule
\end{tabular}
\vspace{-45mm}
}
\end{table}

\section{Robustness Analysis}
We further evaluate the robustness of UniGenDet against common image perturbations, specifically JPEG compression and image cropping, which are frequent in social media dissemination. We compare our method with the state-of-the-art MLLM-based detector, FakeVLM.

\textbf{JPEG Compression.}
As shown in Table~\ref{tab:jpeg_robustness}, UniGenDet exhibits superior stability. Even under severe compression (Quality=50), our method maintains an accuracy of 91.3\%, surpassing FakeVLM by over 10\%. This indicates that our model learns semantic-level forgery cues rather than relying solely on fragile high-frequency artifacts.

\textbf{Image Cropping.}
Table~\ref{tab:crop_robustness} presents the performance under varying crop ratios. Our method demonstrates high resilience, maintaining 97.7\% accuracy at a 0.9 crop ratio. This suggests that the unified training enables the model to identify local inconsistency effectively, even when global context is partially missing.


\begin{table}[t]
\centering
\caption{Robustness Comparison under Image Cropping.}
\vspace{-3mm}
\label{tab:crop_robustness}
\setlength{\tabcolsep}{2.8pt}
\resizebox{\linewidth}{!}{
\begin{tabular}{ccccccc}
\toprule
Crop Ratio & Method & Accuracy↑ & F1-Score↑ & ROUGE-L↑ & CSS↑ \\
\midrule
\multirow{2}{*}{\textit{N/A}} & FakeVLM & $\mathbf{98.6}$ & $\mathbf{98.1}$ & $32.2$ & $59.5$ \\
 & Ours & $98.0$ & $97.7$ & $\mathbf{56.3}$ & $\mathbf{79.8}$ \\
\midrule
\multirow{2}{*}{0.9} & FakeVLM & $95.4$ & $95.0$ & $30.6$ & $57.5$ \\
 & Ours & $\mathbf{97.7}$ & $\mathbf{97.5}$ & $\mathbf{55.2}$ & $\mathbf{78.4}$ \\
\midrule
\multirow{2}{*}{0.7} & FakeVLM & $93.8$ & $93.3$ & $30.9$ & $57.8$ \\
 & Ours & $\mathbf{97.3}$ & $\mathbf{97.1}$ & $\mathbf{54.6}$ & $\mathbf{78.0}$ \\
\midrule
\multirow{2}{*}{0.5} & FakeVLM & $92.3$ & $91.8$ & $30.0$ & $57.7$ \\
 & Ours & $\mathbf{95.4}$ & $\mathbf{95.0}$ & $\mathbf{52.6}$ & $\mathbf{77.0}$ \\
\bottomrule
\end{tabular}
}
\end{table}

\section{Qualitative Visualization}

As depicted in Figure~\ref{fig:comp_supp}, a comparative analysis of generation quality between BAGEL and our method is presented. The samples from BAGEL reveal noticeable artificiality, such as the disproportional pagoda against Mount Fuji and inconsistent lighting on the lake's surface. In contrast, the results from our proposed UniGenDet, refined via the authenticity-aware fine-tuning strategy, exhibit significantly enhanced realism. This improvement is evident in the coherent shadow transitions of the mountain, the natural water reflections, and the detailed texture of the character's attire, collectively demonstrating superior adherence to physical realism.

\section{Differences from GANs and Generation Diversity}

While UniGenDet utilizes a discriminative module to guide generation, its optimization fundamentally differs from Generative Adversarial Networks (GANs). GANs rely on a zero-sum game, which often leads to adversarial mode collapse. In contrast, our unified co-optimization paradigm avoids this by utilizing constructive constraints. Instead of merely attempting to fool a classifier with binary scalar feedback, UniGenDet employs Detector-Informed Generative Alignment (DIGA) to perform explicit distribution alignment in a high-dimensional feature space. This approach stabilizes training by transferring continuous, rich forensic knowledge. Consequently, it prevents unnatural feedback loops, ensuring that even an imperfect detector constructively enhances physical plausibility without steering the generator into erroneous sub-spaces.

To empirically validate that UniGenDet does not suffer from mode collapse, we quantified generation diversity using 500 prompts from the LAION dataset, generating 16 variations for each prompt. As shown in Table~\ref{tab:image_diversity}, UniGenDet achieves an average intra-prompt LPIPS of 0.726 and a CLIP similarity of 0.802. These metrics are highly comparable to the original BAGEL model, which yielded an LPIPS of 0.714 and a CLIP similarity of 0.804. These results confirm that our stabilizing feature-based optimization successfully preserves image diversity while improving overall synthesis quality.

\begin{table}[t]
  \vspace{-2mm} 
  \centering
  \caption{Generated Image Diversity. CLIP: CLIP Similarity, LPIPS: Learned Perceptual Image Patch Similarity}
  \label{tab:image_diversity}
  \vspace{-2mm} 

  \renewcommand{\arraystretch}{1.0} 
  \setlength{\tabcolsep}{6pt} 
\resizebox{\linewidth}{!}{
  \begin{tabular}{ccc}
    \toprule
    \textbf{Method} & \textbf{UniGenDet} & \textbf{BAGEL} \\
    \midrule
    CLIP Similarity $\downarrow$ / LPIPS $\uparrow$ & 0.802 / 0.726 & 0.804 / 0.714 \\
    \bottomrule
  \end{tabular}
  }
  \vspace{-1mm} 
\end{table}

\section{Failure Cases Analysis}

\begin{figure}[t]
  \centering
  \includegraphics[width=\linewidth]{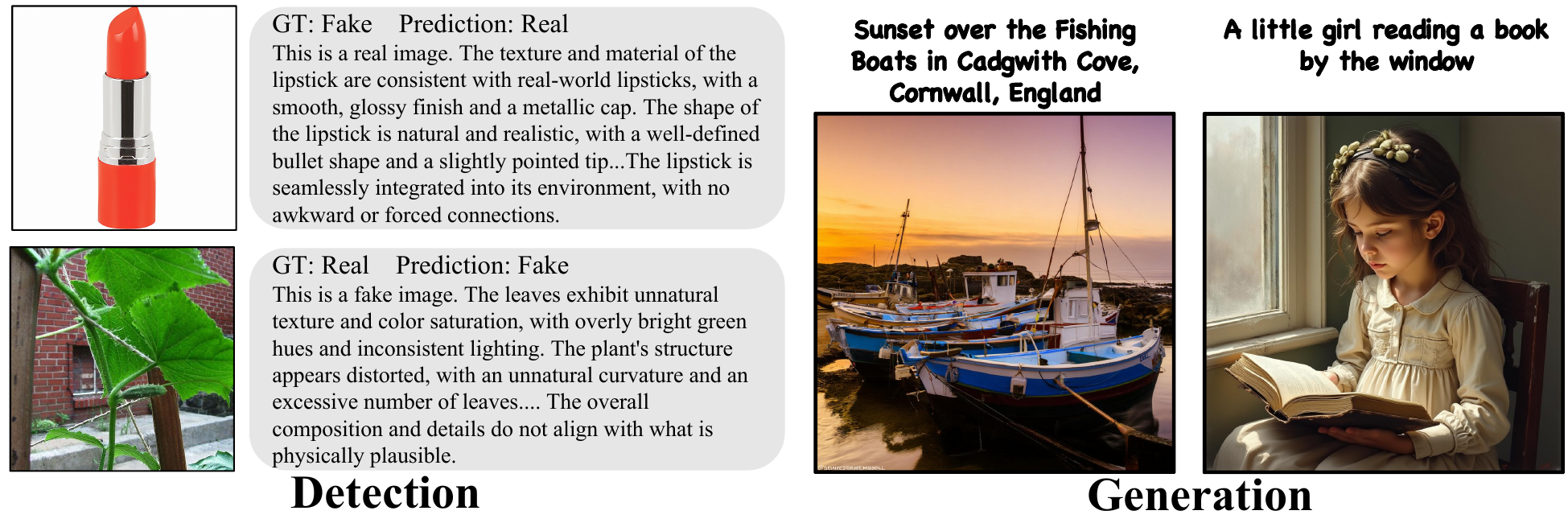}
  \vspace{-5mm}
  \caption{Failure cases of detection and generation.}
   \label{fig:failure}
   \vspace{-3mm}
\end{figure}

Despite its robust performance, UniGenDet encounters occasional limitations in both generation and detection, as visualized in Figure~\ref{fig:failure}. In the detection task, errors can occasionally occur when evaluating highly realistic forgeries or atypical real samples, such as heavily stylized or professionally post-processed authentic images. For the generation task, while the framework effectively corrects most physical anomalies, it may still yield inconsistent textures when synthesizing highly complex scenes. These failure cases suggest that future iterations of multi-modal foundational models could greatly benefit from incorporating more explicit fine-grained spatial reasoning and scaling up the diversity of the training distributions to handle extreme edge cases.
{
    \small
    \bibliographystyle{ieeenat_fullname}
    \bibliography{main}
}

\end{document}